\documentclass[letterpaper, 10 pt, conference]{ieeeconf}  

\IEEEoverridecommandlockouts                             
\usepackage{cite}
\usepackage{caption}

\usepackage{amsmath}
\usepackage{amssymb}
\usepackage{graphicx}
\usepackage{subcaption} 

\usepackage{booktabs}
\usepackage{multirow}
\usepackage{boldline}

\usepackage[ruled,vlined]{algorithm2e}

\usepackage{float}

\title{Bridging VLM and KMP: Enabling Fine-grained robotic manipulation \\via Semantic Keypoints Representation}

\author{Junjie Zhu$^{*1,2,4}$, Huayu Liu$^{*1,2,4}$, Jin Wang$^\dag$$^{1,2,4}$, Bangrong Wen$^{1,2,4}$,\\Kaixiang Huang$^{1,2,4}$, Xiaofei Li$^{1,2,4}$, Haiyun Zhang$^{3,4}$, and Guodong Lu$^{1,2,4}$
\thanks{This paper is supported by the "Pioneer" and "Leading Goose" R\&D Program of Zhejiang Province(2025C01091, 2023C01070), Ningbo Science and technology project(2024Z295).} 
\thanks{ $^*$Indicates equal contribution. }
\thanks{$^\dagger$Corresponding author: {\tt\small dwjcom@zju.edu.cn}.}
\thanks{
	$^{1}$The State Key Laboratory of Fluid Power and Mechatronic Systems, School of Mechanical Engineering, Zhejiang University, Hangzhou 310058, China. $^{2}$Zhejiang Key Laboratory of Industrial Big Data and Robot Intelligent Systems, Zhejiang University, Hangzhou 310058, China. $^{3}$School of Robotics, Ningbo University of Technology, Ningbo 315211, China. $^{4}$Robotics Research Center of Yuyao City, Ningbo 315400, China.   }
}

\begin{document}

\maketitle
\thispagestyle{empty}
\pagestyle{empty}

\begin{abstract}
From early Movement Primitive (MP) techniques to modern Vision-Language Models (VLMs), autonomous manipulation has remained a pivotal topic in robotics. As two extremes, VLM-based methods emphasize zero-shot and adaptive manipulation but struggle with fine-grained planning. In contrast, MP-based approaches excel in precise trajectory generalization but lack decision-making ability. To leverage the strengths of the two frameworks, we propose VL-MP, which integrates VLM with Kernelized Movement Primitives (KMP) via a low-distortion decision information transfer bridge, enabling fine-grained robotic manipulation under ambiguous situations. One key of VL-MP is the accurate representation of task decision parameters through semantic keypoints constraints, leading to more precise task parameter generation. Additionally, we introduce a local trajectory feature-enhanced KMP to support VL-MP, thereby achieving shape preservation for complex trajectories. Extensive experiments conducted in complex real-world environments validate the effectiveness of VL-MP for adaptive and fine-grained manipulation.
\end{abstract}

\section{Introduction}

Autonomous manipulation has consistently remained a central and enduring research topic in robotics\cite{1}. From MP-based methods prior to the advent of large language models to the current VLM-based approaches, this field has experienced a significant paradigm shift and technological evolution.

Although traditional imitation learning has seen extensive development, the introduction of MLLMs/VLMs has divided contemporary robotic manipulation methods into two distinct extremes: (1) The end-to-end VLM-based agent, which implicitly models robot manipulation and directly outputs actions \cite{2,3}, and (2) The traditional MP-based methods in imitation learning\cite{20},\cite{21}. Today, mainstream methods leverage the adaptive perception of VLMs, focusing on skill generalization and decision-making capabilities. Although these methods have achieved skill generalization beyond traditional imitation learning, their precision of manipulation and inference efficiency remain unguaranteed \cite{4,5}. In contrast, MP-based methods excel in the precise generalization of skill trajectories. Therefore, a natural thought is bridging the gap between these two extremes to achieve fine-grained manipulation with task parameter representation.

\begin{figure}
    \centering
    \captionsetup{font={footnotesize }}
    \includegraphics[width=1\linewidth]{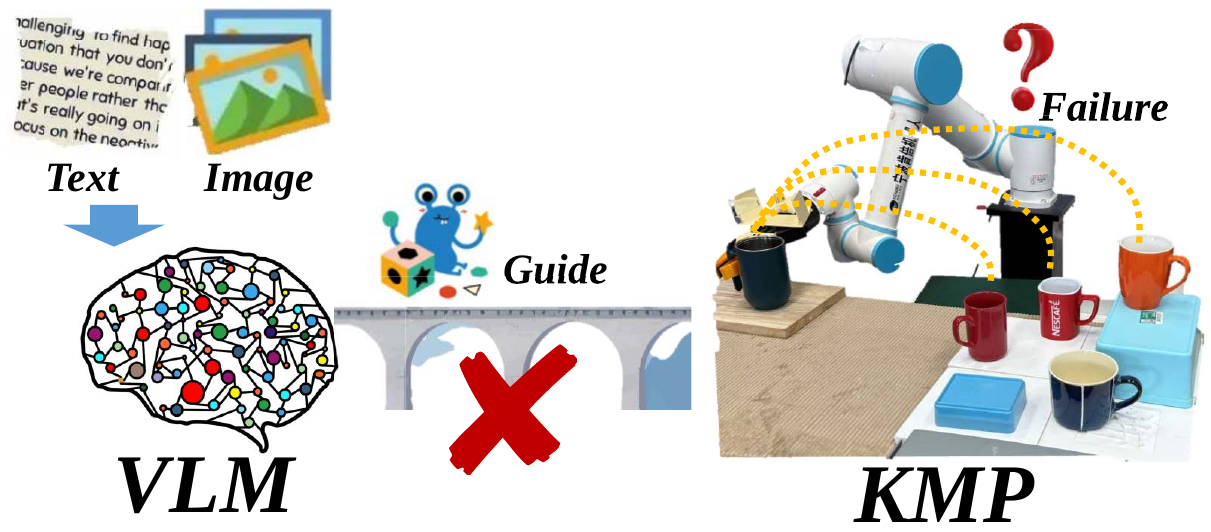}
    \caption{Illustration of the dilemma of current robotic manipulation, without the precise guidance of VLM decision information, KMP methods struggle to generalize fine-grained tasks, such as pouring water into a specified cup. }
    \label{fig:Intro}
\end{figure}

Unfortunately, as shown in Fig. \ref{fig:Intro}, there is limited research in bridging VLM and KMP methods, while the decision information expressed by VLM is not fully utilized in the lower-level trajectory generalization processes. Few studies focus solely on action function calls generated by LLM \& VLM, such as ChatGPT for Robotics \cite{6}, which explores the code generation capabilities of LLM to guide robot action function calls. Liang \cite{7} further proposed a robot-centric formalization of LLM-generated programs that can represent reactive policies. However, these studies primarily focus on task sequence decisions and neglect the parameter transfer process for specific task trajectory generalization, leading to poor fine-grained manipulation. These limitations mainly stem from the following challenges: (i) Information distortion occurs when VLM decision information is transformed into MP task parameters, resulting in the insufficient utilization of prior decision information from VLM by lower-level trajectory generalization algorithms; (ii) Current MP-based imitation learning methods overlook the local features near the start and end points of the demonstration trajectory during trajectory generalization, causing distortion in the generated trajectory’s shape, ultimately leading to failed fine-grained manipulation.

To address these challenges, this paper proposes VL-MP, a novel framework that bridges VLM and KMP in robot manipulation through semantic keypoint constraints and local trajectory feature enhancement. VL-MP effectively represents task generalization parameters by extracting and constraining 3D semantic keypoints from the decision information provided by VLM. The local feature-enhanced KMP (LFE-KMP) is then employed to generalize skill trajectories that preserve their original shape, enabling autonomous decision-making and fine-grained manipulation in ambiguous environments. We conduct extensive real-world experiments in scenarios involving pouring tasks and further quantify its ability to maintain complex trajectory shapes through simulation experiments.

In summary, our main contributions are threefold:

\begin{enumerate}
    \item We introduce VL-MP, a novel robot manipulation generalization framework that effectively bridges the decision-making capability of VLM with the generalization ability of the KMP method, enabling autonomous decision-making and fine-grained manipulation generalization in the real-world environment.
    \item The Bridge Layer is proposed for task parameter transfer in VL-MP, which employs 3D semantic keypoint constraints to accurately capture the decision results of the VLM. This provides a target shape description for lower-level skill generalization, addressing the insufficient utilization of decision information, leading to fine-grained manipulation from the source.
    \item   The local trajectory feature enhanced KMP is proposed, which emphasizes  manipulation details around the start and end points of the demonstration trajectory. This approach ensures the overall trajectory shape without distortion, significantly improving the success rate of fine-grained manipulation generalization.
\end{enumerate}

\section{Related Works}

\subsection{VLM for Manipulation}

Harnessing VLMs \cite{8} to steer robotic manipulation remains a dynamic avenue of research. While notable advancements have been achieved in open-world decision-making and target localization\cite{9},  the caption-driven pre-training paradigm of VLMs inherently constrains the retention of intricate visual details within images \cite{11}, thereby impeding the multimodal reasoning capabilities of robotic systems. Conversely, self-supervised visual models such as DINO \cite{12} offer fine-grained, pixel-level representations that prove advantageous for a wide range of vision and robotics applications \cite{13}, yet they falter in grasping open-world semantics essential for instructional reasoning—a fundamental aspect of goal-driven decision-making in ambiguous environments. In this study, we address these challenges by coupling GPT-4o \cite{27} with GroundingDINO\cite{14} for coupled visual reasoning and instruction generation, leveraging their complementary strengths for task decision-making and downward transfer through the Bridge Layer.

\subsection{Bridge for Task Parameter Transfer}

The bridge for task parameter transfer determines the coordination method between different modules in the robotic system and exerts distinct impacts on manipulation precision and success rates. Many studies adopt data-driven approaches to investigate task parameter transfer, such as learning object-centric representations\cite{15} or scene descriptors\cite{17}, which transfer static coordinates through training. Some research directly employs VLM to output 3D voxel maps for motion planning models \cite{18,19}. However, these methods exhibit poor interpretability at the trajectory generalization level, fail to generalize across subtle variations within object instances, and lack informational support for fine-grained manipulations. Robotic manipulations often involve intra-class object instances with diverse shape characteristics, necessitating a robust bridge layer to encode decision-making information into task parameters that guide trajectory generalization precisely. Here, we design the Bridge Layer that characterizes the geometric features of the target via semantic keypoints and constrains the solution to obtain precise task parameters, thereby ensuring fine-grained accuracy and shape preservation during the trajectory generalization.

\subsection{Movement Primitives for Trajectory Learning}

Movement Primitives (MP) are a mathematical framework used in robotics and machine learning to model and reproduce complex joint behaviors. Dynamic Movement Primitives (DMP)\cite{20} is a method for modeling robotic behaviors, demonstrating its flexibility, adaptability, and potential for integration with statistical learning. After that, an enhanced MP method termed Kernelized Movement Primitives (KMP)\cite{21} was introduced. This method allows for the activation of movement primitives through a combination of mixed actions, task description parameters, and probability functions, thereby inheriting the advantages of probability function and addressing limitations in DMP and Probabilistic Movement Primitives (ProMP)\cite{22}, such as the need to specify the form and number of basis functions explicitly. Although KMP exhibits strong spatial trajectory reproducibility, it has weak robustness to disturbances and poor generalization performance, often resulting in trajectories with severe shape distortion, leading to failure in autonomous manipulation. In this study, based on KMP, we propose LFE-KMP in our VL-MP framework, which enhances local trajectory features near task parameters through localized sampling, eliminates dependency on absolute trajectory positions, and improves generalization and stability. 

\section{Methods}

Here we discuss:

\begin{enumerate}
    \item  The overall architecture of VL-MP.
    \item How to construct a bridge layer based on Semantic Keypoints Constraints to extract precise task parameters from VLM decision-making information and ensure their robust downstream transmission.
    \item How to generate fine-grained manipulation trajectories using Local Feature Enhanced KMP, achieving one-shot trajectory generalization with minimal shape distortion. 
\end{enumerate}

\begin{figure*}[t!]
	\centering
	\includegraphics[width=1\textwidth]{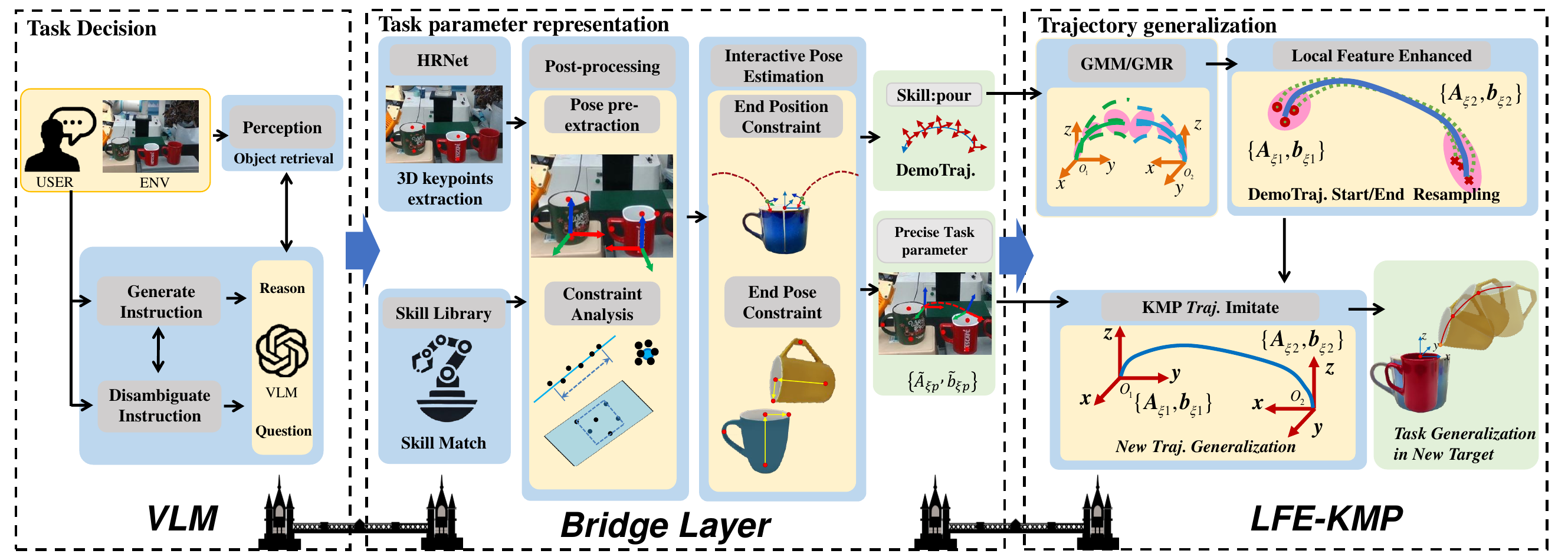} %
        \captionsetup{font={footnotesize }}
	\caption{Overview of VL-MP. In an ambiguous task, we first employ the VLM to process environmental and language inputs for task decision-making. Subsequently, the proposed Bridge Layer extracts 3D keypoints and provides an accurate representation of the task parameters, which are then transmitted downward without distortion. Finally, at the LFE-KMP stage, an accurate generalization of one-shot tasks is achieved through KMP modeling by resampling the local features of the demonstration trajectory.  }
	\label{fig:Framework of VL-MP}
\end{figure*}

\subsection{ Method Overview }

The framework diagram(Fig. \ref{fig:Framework of VL-MP}) illustrates the pipeline of the proposed VL-MP. We employ a bridge layer to process and re-represent the task decision information from the VLM for the LFE-KMP to generalize new task trajectories, significantly improving the precision of task parameter representation. In detail, during the VLM-based perception and decision-making stage, we utilize multiple VLM program blocks for interaction to construct a robot decision-making framework based on multi-turn dialogue. 
The system accomplishes object-centric goal determination through collaborative interactions among three modules: Generate Instruction (GI), Disambiguate Instruction (DI), and Perception. Subsequently, the bridge layer utilizes HRNet for keypoint detection, evaluates task constraints based on variance distributions of interactive poses in the skill library, and refines them through topological relationships among 3D keypoints. The poses of the source and target objects are derived from keypoint-based reconstruction. The source pose establishes the initial task parameters, while the target object's pose dictates positional and orientational constraints. By employing position and pose sampling, the terminal pose parameters of the trajectory are systematically generated, serving as new task inputs\(\{\hat{\text{A}}_{\xi p}, \hat{\text{b}}_{\xi p}\}\) for the LFE-KMP. Finally,  guided by the trajectory’s start/end reference points, the LFE-KMP generalizes fine-grained manipulation trajectories. 

\begin{figure}
    \centering
    \captionsetup{font={footnotesize }}
    \includegraphics[width=1\linewidth]{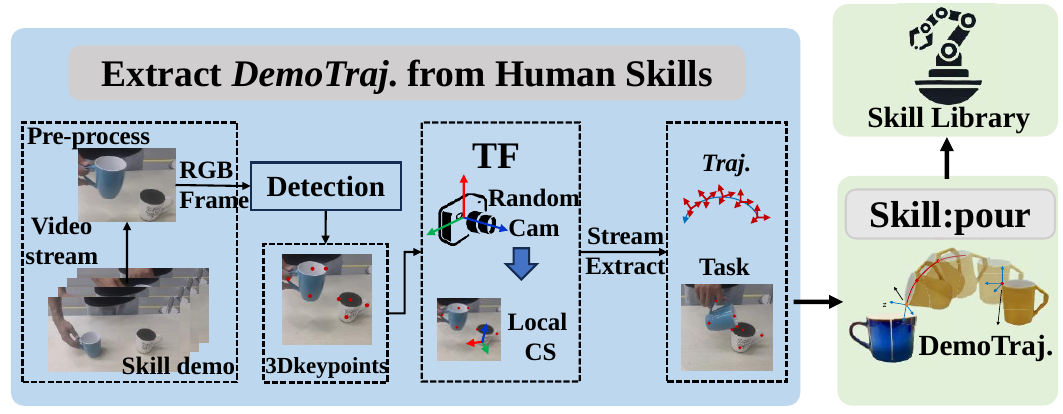}
    \caption{Illustration of $[DemoTraj.]$ extraction. Keypoints are extracted from the skill video stream. In each frame, a local coordinate system is constructed based on the keypoints of the manipulated object to capture its pose. The last frame detection result is used as the interaction state of the skill.}
    \label{fig:DemoTraj}
\end{figure}

\subsection{Task Parameter Representation via VL-MP}
In current robotic manipulation frameworks, the representation and transmission of decision-making information are severely distorted. This results in insufficient trajectory accuracy, which frequently leads to collisions. To address this issue, we propose the VL-MP manipulation framework and design the Bridge Layer based on 3D semantic keypoint representation and constraints.  

\begin{figure}
    \centering
    \captionsetup{font={footnotesize}}
    \includegraphics[width=1\linewidth]{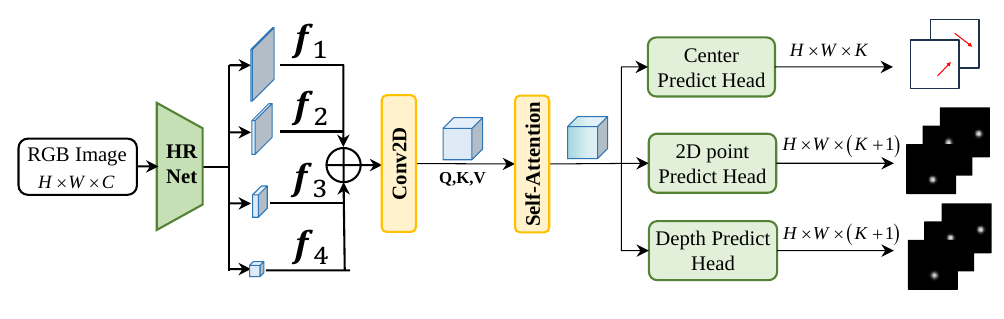}
    \caption{The diagram of keypoints detection network. The keypoints detection network utilizes HRNet for multi-scale feature extraction from images. Self-attention is then applied to further enhance the global dependencies of the features. Finally, three output heads are employed to predict the center points, keypoints, and depth.}
    \label{fig:HRNet}
\end{figure}

\begin{figure*}[htbp]
    \centering
    \captionsetup{font={footnotesize }}
    \begin{subfigure}[t]{0.18\linewidth} 
        \centering
        \includegraphics[width=\linewidth]{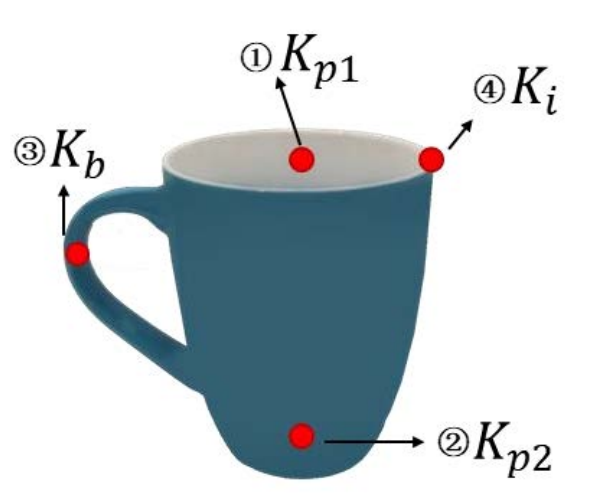}
        \caption{Diagram of 3D keypoints}
        \label{fig:Diagram-of-3D-keypoints}
    \end{subfigure}
    \hfill
    \begin{subfigure}[t]{0.18\linewidth} 
        \centering
        \includegraphics[width=\linewidth]{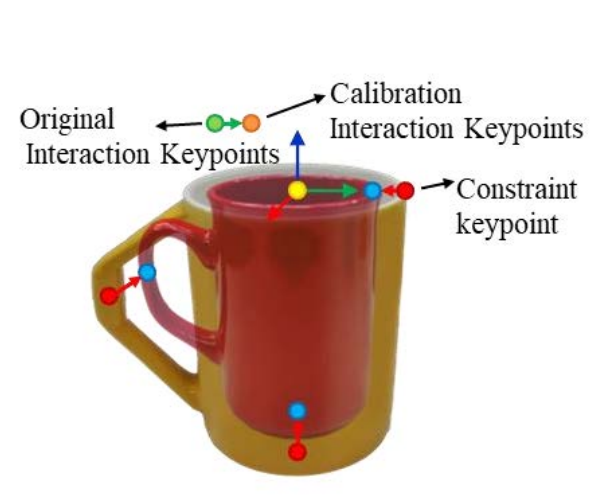}
        \caption{Keypoints Alignment}
        \label{fig:Keypoints-Alignment}
    \end{subfigure}
    \hfill
    \hfill
    \begin{subfigure}[t]{0.18\linewidth} 
        \centering
        \includegraphics[width=\linewidth]{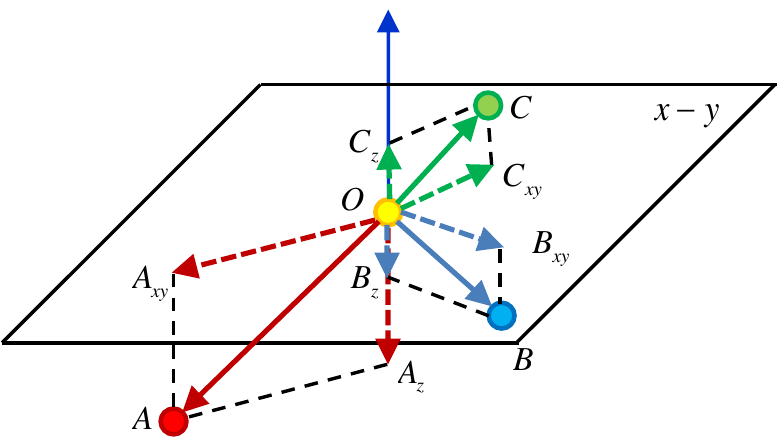}
        \caption{Keypoints Proj. on x-y Plane}
        \label{fig:Keypoints-Proj.}
    \end{subfigure}
    \hfill
    \begin{subfigure}[t]{0.18\linewidth} 
        \centering
        \includegraphics[width=\linewidth]{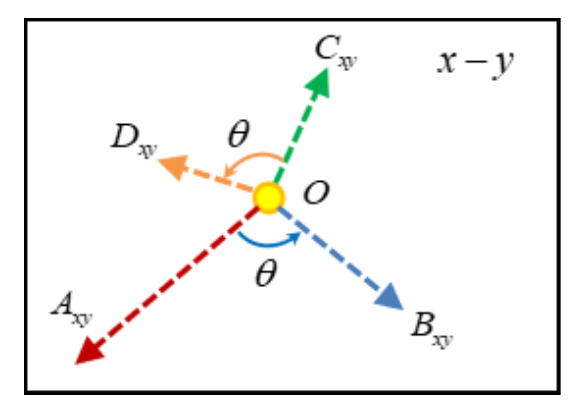}
        \caption{Modification of Interaction Keypoints}
        \label{fig:Modification-of-Interaction-Keypoints}
    \end{subfigure}
        \begin{subfigure}[t]{0.18\linewidth} 
        \centering
        \includegraphics[width=\linewidth]{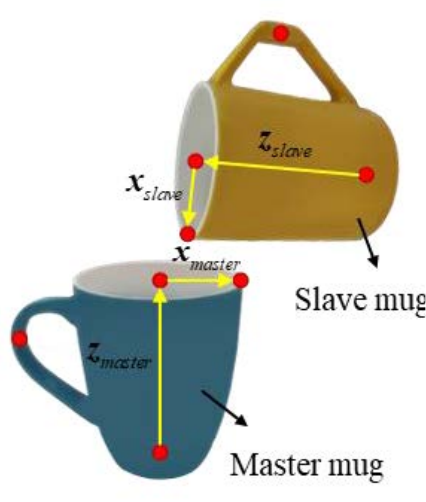}
        \caption{EndPose-Estimate}
        \label{fig:EndPose-Estimate}
    \end{subfigure}
    \caption{Illustration of the keypoints definition and task constraint. (a) Describes the definition of 3D keypoints. (b) Demonstrates keypoint normalization and calibration across instances within the same class, aimed at more accurate task parameter representation. (c) Describes the construction of the termination task posture, used for subsequent sampling and estimation of the termination task pose. (d) (e) Illustrate the derivation process of the normalized calibration interaction keypoints through the x-y plane projection of keypoints. }
    \label{fig:keypoint-define&analysis}
\end{figure*}

In the 3D keypoint representation stage, to accommodate environment variability, we concisely define three categories of keypoints—interaction keypoints, positional keypoints, and boundary keypoints (Fig. \ref{fig:Diagram-of-3D-keypoints})—to describe interactive details (e.g., target scale, shape, and orientation) and geometric constraints during manipulation tasks. HRNet\cite{23} is used for keypoint generation(Fig. \ref{fig:HRNet}), as exemplified by the mug in water-pouring tasks where four keypoints are generated: an interaction keypoint \( K_i \) (rim), two positional keypoints  \( K_p \) (cup mouth center and base center), and a boundary keypoint  \( K_b \) (handle). The VLM decision output ${x}\in{R}^{{H}\times{W}\times{C}}$ feeds into HRNet's backbone network to extract four multi-scale feature maps \( f_1 \in \mathbb{R}^{H \times W \times C} \), \( f_2 \in \mathbb{R}^{\frac{H}{2} \times \frac{W}{2} \times 2C} \), \( f_3 \in \mathbb{R}^{\frac{H}{4} \times \frac{W}{4} \times 4C} \), and \( f_4 \in \mathbb{R}^{\frac{H}{8} \times \frac{W}{8} \times 8C} \).  These features undergo unified upsampling followed by Conv2D-based multi-scale fusion, then pass through a self-attention module to expand the receptive field and capture global dependencies, yielding enhanced fused features  \( F \in \mathbb{R}^{H \times W \times C} \). Three prediction heads subsequently process these features to estimate: (i) center coordinates,(ii) 2D keypoint positions $({u},{v})$, and (iii) keypoint depth information ${z}$.

\begin{figure*}
\begin{equation}
\overrightarrow{OD}_{xy} = |\overrightarrow{OC}_{xy}| \frac{|\overrightarrow{OB}_{xy}| \cos(\theta) \cdot \overrightarrow{OA}_{xy} + \sin(\theta) \cdot (\overrightarrow{z} \times \overrightarrow{OA}_{xy}) + (1 - \cos(\theta))(\overrightarrow{z} \cdot \overrightarrow{OA}_{xy})\overrightarrow{z}}{|\overrightarrow{OA}_{xy}| \cos(\theta) \cdot \overrightarrow{OA}_{xy} + \sin(\theta) \cdot (\overrightarrow{z} \times \overrightarrow{OA}_{xy}) + (1 - \cos(\theta))(\overrightarrow{z} \cdot \overrightarrow{OA}_{xy})\overrightarrow{z}}
\label{eq:f4}
\end{equation}
\end{figure*}

Finally, the semantic keypoints \(\ K = ( K_1, K_{p1}, K_{p2}, K_b )\)  can be computed using the following formula(\ref{eq:f1}), where \( K_{cam} \) is the camera intrinsic matrix, while \( R \) and \( t \) represent the rotation matrix and translation vector in the camera extrinsic matrix, respectively. 
 \begin{equation}
K = R \cdot \left( z \cdot K_{cam}^{-1} - 1 \cdot \mathbf{ \begin{bmatrix} u \\ v \\ 1 \end{bmatrix} } \right) + t
\label{eq:f1}
\end{equation}

Within the skill library(Fig. \ref{fig:DemoTraj}), the interaction state information derived from the final frames of multiple demonstrations plays a pivotal role in ensuring successful skill execution, as the shape, size, and orientation of the target object impose constraints on the manipulated object. By assimilating these interaction states and representing the target object's condition through 3D keypoints, the terminal pose of the task trajectory can be deduced, thereby furnishing precise termination parameters \( P_{target} \) for the LFE-KMP algorithm.

To learn the interaction states in demonstrations, we constrain the termination task parameters ${pos}$ using the positional distribution of the interaction keypoints \( K_i \) of the manipulated object. Since variations in the shape and size of the target object in demonstrations can influence  \( K_i \), we extract the final frame of all demonstrations and normalize the termination position ${pos}$ by selecting the target object's all the keypoints except \( K_{p1} \) for correction.

\( K_i \) is considered to be the primary constraint on the interaction keypoints of the manipulated object(slave mug in Fig.\ref{fig:EndPose-Estimate}). As illustrated in Fig.\ref{fig:Keypoints-Alignment}, a constant topological relationship (\ref{eq:f2}) is constructed during the normalization. Before and after displacement, as shown in Fig. \ref{fig:Keypoints-Proj.} and  \ref{fig:Modification-of-Interaction-Keypoints}, it is denoted as A and B, respectively. \( K_{p1} \) corresponds to O, while the interaction keypoint is C before correction and D after correction. This relationship yields the corrected interaction keypoint \( K_i \) (\textit{A} in Fig. \ref{fig:Keypoints-Proj.} and  \ref{fig:Modification-of-Interaction-Keypoints}),  \(pos\) (the coordinates of D) can be calculated via the following formula(\ref{eq:f3},\ref{eq:f4}):

\begin{equation}
\left\{
\begin{aligned}
\langle \overrightarrow{OA_{xy}}, \overrightarrow{OB_{xy}} \rangle &= \langle \overrightarrow{OC_{xy}}, \overrightarrow{OD_{xy}} \rangle = \theta \\
\frac{|\overrightarrow{OB_{xy}}|}{|\overrightarrow{OA_{xy}}|} &= \frac{|\overrightarrow{OD_{xy}}|}{|\overrightarrow{OC_{xy}}|}
\end{aligned}
\right.
\label{eq:f2}
\end{equation}

\begin{equation}
\text{pos} = \overrightarrow{0} + \overrightarrow{OD}_{xy}
\label{eq:f3}
\end{equation}

The pose distribution of the manipulated object constrains the termination task pose. We select  \( K_i \) to construct the orientation vector ${Z}$, where ${a}$ is defined as the angle between the vector of the target interaction object \( Z_{master} \) and that of the manipulated object \( Z_{slave} \)(Fig. \ref{fig:EndPose-Estimate}). Finally, sampling(Algorithm\ref{alg:EndposeEstimate}) is performed based on the variance distributions of ${pos}$ and ${a}$, aiding the Bridge Layer in task parameter inference. The skill termination pose and initial pose collectively form new task parameters $\{\hat{\text{A}}_{\xi p}, \hat{\text{b}}_{\xi p}\}_{p=1}^P$ for subsequent trajectory generalization. 

In summary, within the VL-MP framework, we implement secondary refinement of VLM-derived decision information through keypoint detection and further generate precise task termination parameters based on keypoint constraints, thereby establishing the foundation for subsequent accurate trajectory generalization. 

\begin{algorithm}
\SetAlgoLined
\SetKwInput{Input}{Input}
\SetKwInput{Output}{Output}
\SetKwInput{Initialization}{Initialization}
\footnotesize
\caption{EndposeEstimate}
\label{alg:EndposeEstimate}
\Input{Keypoint coordinate sets \(U_{slave}\) and \(U_{master}\), termination position \(\boldsymbol{pos}\), and termination orientation angle \(\alpha\)}
\Output{Skill termination pose \(P_{target}\)}
\Initialization{Initial pose of the manipulated object \(P_{start}\)}
Sample interaction keypoints based on the termination position \(\boldsymbol{pos}\) to obtain \(u^{target}\)\;
\For{\(u\) in \(U_{master}\)}{
    \If{\(u\) in \(U_{master}^{constraint}\)}{
        Normalize and correct \(u^{target}\)\;
    }
}
Express \(u^{target}\) in the original coordinate system\;
With \(\alpha\) under constraint, sample a series of candidates \(z_{slave}\) at intervals of \(2^{\circ}\), and calculate the pose distance with \(P_{start}\)\;
Take the pose with the smallest distance relative to \(P_{start}\) as the target pose, and combine it with \(u^{target}\) to obtain \(P_{target}\)\;
\Return \(P_{target}\); 
\end{algorithm}

\subsection{Trajectory Generalization via VL-MP}

The KMP algorithm struggles to capture subtle variations near the trajectory's start and end points, leading to limitations in generalizing fine-grained trajectories while preserving their shape. In this part, we propose LFE-KMP by incorporating local trajectory feature augmentation, thereby improving its generalization capability. 

In the Local Feature Enhanced KMP part, we define the pose trajectory sequence extracted from videos as the Demonstration Trajectory \(\xi(t) = \{s, q\}\), comprising both position \(s = [x, y, z]^T\) and orientation \(q = [q_1, q_2, q_3, q_4]^T\). During the preprocessing stage, we represent the demonstration trajectory on a Riemannian manifold to strengthen the constraint properties of orientation. While KMP approaches employ GMM/GMR to model and regress demonstration trajectories for obtaining reference trajectory distributions, the robustness of subsequent KMP models degrades when execution regions significantly differ from demonstration regions. To address this and better capture local motion details, we encode the demonstration trajectory in task-parameterized local coordinate systems, which are intrinsically linked to the task.  Subsequently, under varying task parameters, the  GMM/GMR algorithm is employed to model and regress the probabilistic distribution $\hat{\xi}(t) \sim \mathcal{N}(s,q|\hat{\mu}_s, \hat{\boldsymbol{\Sigma}}_s,\hat{\mu}_q,\hat{\boldsymbol{\Sigma}}_q)$ of the trajectory, serving as the skill reference trajectory for each task parameter.

Following the original KMP, we use the reference trajectory obtained from GMR regression as a basis and apply a kernel function to re-fit the probability distribution of the demonstration trajectory, denoted as $\xi(t)$. We learn the parameterized trajectory by minimizing the Kullback-Leibler (KL) divergence between it $\xi(t)$ and the reference trajectory $\hat{\xi}(t) \sim \mathcal{N}(s,q|\hat{\mu}_s,\hat{\boldsymbol{\Sigma}}_s,\hat{\mu}_q,\hat{\boldsymbol{\Sigma}}_q)$. This optimization ensures minimal information loss during the imitation learning process.

 However, due to the interactions among multiple task parameters, the skill trajectory may struggle to converge to the desired start/end poses. To address this, we propose a task-parameterized local feature enhancement method to improve the KMP. Specifically, for each task parameter, the algorithm locally resamples the start and end points of the modeled reference trajectory $\hat{\xi}$ obtained through regression. These sampled points are treated as desired trajectory points $P_e$, with covariance set to $\boldsymbol{\Sigma} = \varepsilon \boldsymbol{I}^{o}$, where $\varepsilon$  is an infinitesimal value. The desired trajectory points $P_e$ are merged with the original reference trajectory $\hat{\xi}$ to form an extended reference trajectory. The KMP algorithm is then applied to predict new mean and covariance values. The KL divergence loss function for the final LFE-KMP, which integrates task-parameterized local feature enhancement and kernel-based fitting, is formulated as \ref{eq:f8}.
\begin{equation}
\underset{\mu_w,\Sigma_w}{\text{argmin}} J = \sum_{i = 1}^{N + R} D_{KL}\left(P_{\mathrm{p}}\left(\boldsymbol{\xi} \vert t_{i}^{\text{extend}}\right) \| P_{\mathrm{r}}\left(\boldsymbol{\xi} \vert t_{i}^{\text{extend}}\right)\right)
\label{eq:f8}
\end{equation}

In this formulation, $t^{\text{extend}}$ denotes the extended time inputs combining the desired points and the original reference trajectory. 

To generalize skill trajectories across new task parameters, the local LFE-KMP results under the current task parameters undergo a Gaussian linear transformation to derive the local LFE-KMP trajectory distribution $\tilde{\xi}(t^*)^{(p)} = \{\tilde{s}(t^*)^{(p)}, \tilde{q}(t^*)^{(p)}\}$ in the global coordinate system. Finally, the generalized global trajectory for new tasks is determined by maximizing the product of Gaussians under the new task parameters $\{\tilde{\boldsymbol{A}}_{\tilde{\zeta},p},\tilde{\boldsymbol{b}}_{\tilde{\zeta},p}\}_{p = 1}^{P}$: 
\begin{equation}
\tilde{\boldsymbol{s}}(t^*)=\underset{\boldsymbol{s}}{\text{argmax}}\prod_{p = 1}^{P} \mathcal{N}\left(\tilde{\boldsymbol{A}}_{\tilde{\zeta}p} \tilde{\boldsymbol{\mu}}_{s}^{(p)}+\tilde{\boldsymbol{b}}_{\tilde{\zeta}p},\tilde{\boldsymbol{A}}_{\tilde{\zeta}p} \tilde{\boldsymbol{\Sigma}}_{s}^{(p)} \tilde{\boldsymbol{A}}_{\tilde{\zeta}p}^{\mathrm{T}}\right)
\label{eq:f9}
\end{equation}
\begin{equation}
\tilde{\boldsymbol{q}}(t^*)=\underset{\boldsymbol{q}}{\text{argmax}}\prod_{p = 1}^{P} \mathcal{N}\left(\tilde{\boldsymbol{\mu}}_{q\_\text{global}}^{(p)},\boldsymbol{\Lambda}^{\mathrm{T}} \tilde{\boldsymbol{\Sigma}}_{q}^{(p)} \boldsymbol{\Lambda}\right)
\label{eq:f10}
\end{equation}
In this formulation:

\[
\tilde{\boldsymbol{\mu}}_{q\_\text{global}}^{(p)} = \mathrm{Exp}_{q_{\tilde{\boldsymbol{A}}_{\tilde{\zeta}p}}}\left(\mathrm{Log}_{e_0} \tilde{\boldsymbol{\mu}}_{q}^{(p)}\right),\boldsymbol{\Lambda} = \Gamma_{q_{\tilde{\boldsymbol{A}}_{\tilde{\zeta}p}} \to \tilde{\boldsymbol{\mu}}_{q\_\text{global}}^{(p)}}\left(\mathbf{I}_3\right)
\]The probability distribution \(\xi_{new}\) of the final trajectory in the global coordinate system is obtained through computation. Subsequently, sampling is performed on this distribution \(\xi_{new}\) to generate the desired task trajectory.

In summary, the proposed LFE-KMP enhancement method effectively improves the trajectory's learning and adaptation capabilities to local variations, enabling the VL-MP framework to achieve precise generalization of task trajectories for complex-shaped objects. 

\begin{figure*}[t!]
	\centering
        \captionsetup{font={footnotesize }}
	\includegraphics[width=1\textwidth]{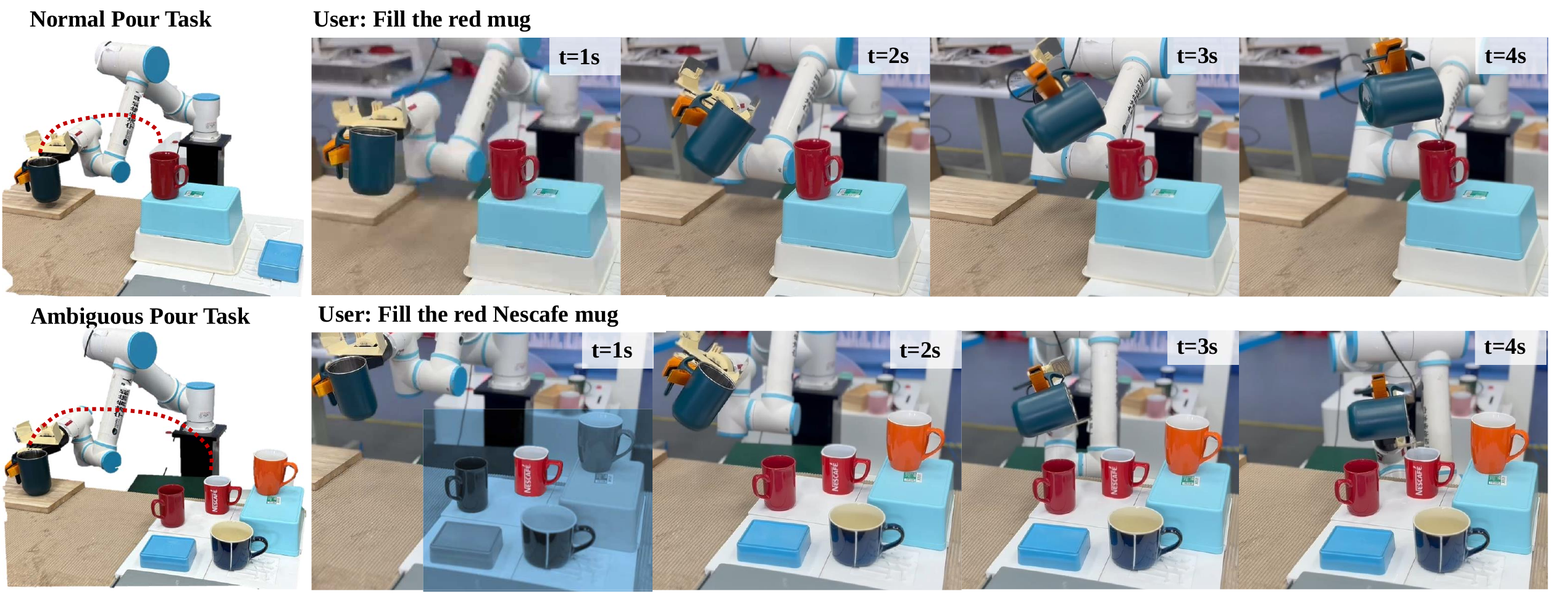} %
    \caption{Illustration of the real-world pouring experiment results for two environments. In the normal task, the task instruction remains fixed, with no decision-making required. In contrast, in the ambiguous task, the task instruction is randomized, necessitating decision information transmission and revalidation of task parameters. The figure presents the task execution outcomes  ${t=1s}$ in both environments, the VL-MP strategy ensures that the pouring process is successfully completed without spillage. }
    \label{fig:Pouring Performance}
\end{figure*}

\section{Experiments and Discussion}

We elaborate on the implementation details of the VL-MP framework. Subsequently, we conduct real-world experiments on water-pouring tasks across diverse environments, validating the effectiveness of fine-grained manipulation through comparative success rate analysis. The error distribution during task execution is systematically analyzed to identify failure sources and propose improvement strategies. Further investigations explore how VL-MP preserves skill trajectory shape during generalization to learn more challenging tasks, with quantitative verification through simulation experiments.
\begin{figure}
    \centering
    \captionsetup{font={footnotesize }}
    \includegraphics[width=1\linewidth]{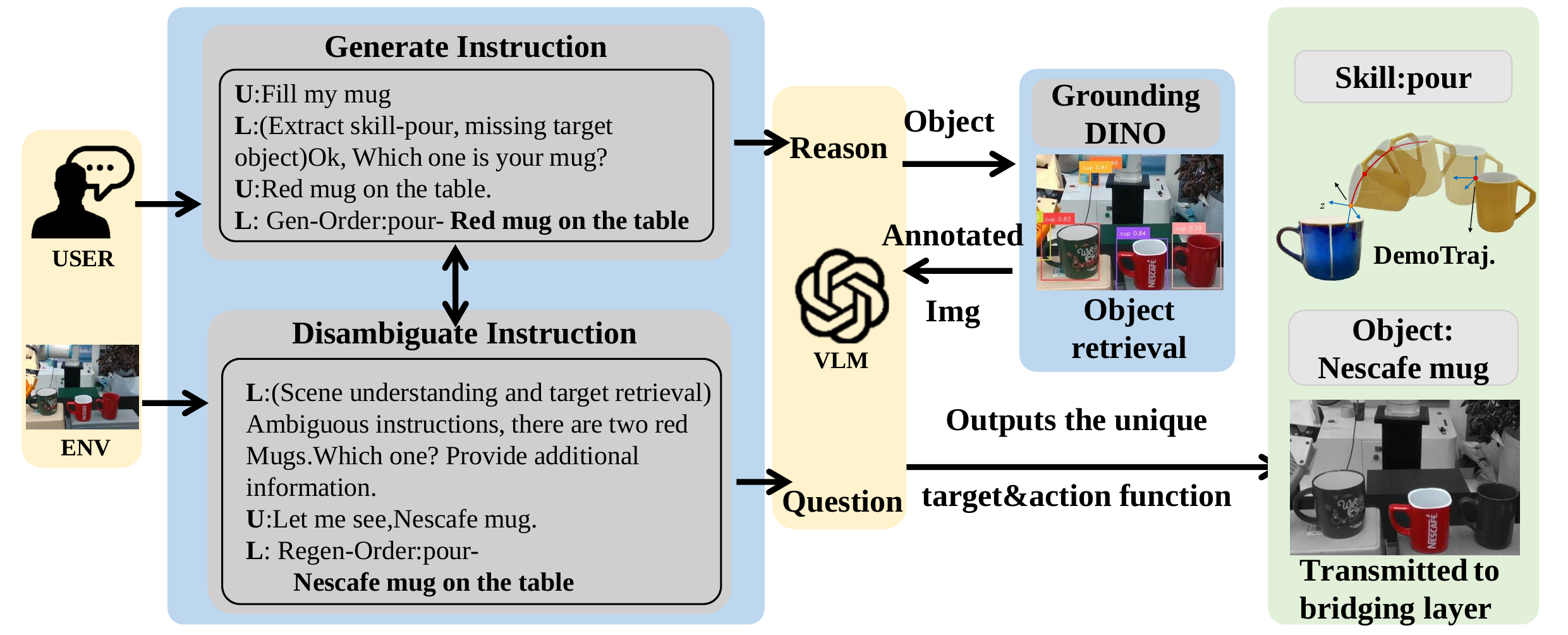}
    \caption{Illustration of VLMs Prompting and Perception. GI processes natural language inputs from users, generating preliminary standardized instructions. Target objects are retrieved via the GroundingDINO to obtain candidate object groups. DI then is activated, sharing historical dialogue context with the GI and leveraging candidate retrieval data.DI  resolves ambiguities and selects a unique target object for task execution.}
    \label{fig:VLM decisions}
\end{figure}

 \textit{\textbf{VLMs Prompting and Perception}}(Fig. \ref{fig:VLM decisions}). We adopt the prompt engineering framework proposed by Liang et al. \cite{7}, which employs recursively self-generated code to invoke the LLM. Each Language Model Program (LMP) is responsible for an independent function (e.g., environmental perception). VL-MP utilizes GPT-4o \cite{27} via the OpenAI API, designing two LMPs: the GI (General Intelligence) and DI (Domain Intelligence) modules. For the DI and GI modules, we specifically designed their system roles and prompts, including background information, skill library specifications, task definitions, input-output controls, and detailed prompt templates. Additionally, 10 exemplar query-response pairs are incorporated as supplementary prompts. In the perception pipeline, GroundingDINO utilizes object descriptions from the LLM module as queries to generate detection masks, which are then fed to the Language-Motion Planner (LMP) for ambiguity resolution.

 \textit{\textbf{Skill Demo-Trajectory Extraction}}(Fig. \ref{fig:DemoTraj}).In this stage, each skill in the skill library is fundamentally characterized by one-shot learning and the generalization of trajectories. We employ the aforementioned keypoint detection network to perform frame-by-frame tracking and extraction from human demonstration videos, thereby acquiring corresponding skill trajectory data. To address the randomness of camera positions during video recording, we construct a local coordinate system based on the manipulated object's keypoints, enabling the capture of its pose variations as skill pose trajectory information, while utilizing the detection results from the final frame as the skill's interaction state representation(utilized by the aforementioned termination task pose estimation algorithm). During the trajectory learning stage, hyperparameter values of the kernel function are automatically optimized through root mean square error minimization.

\subsection{Pouring Task}
\begin{figure}
    \centering
    \captionsetup{font={footnotesize }}
    \includegraphics[width=1\linewidth]{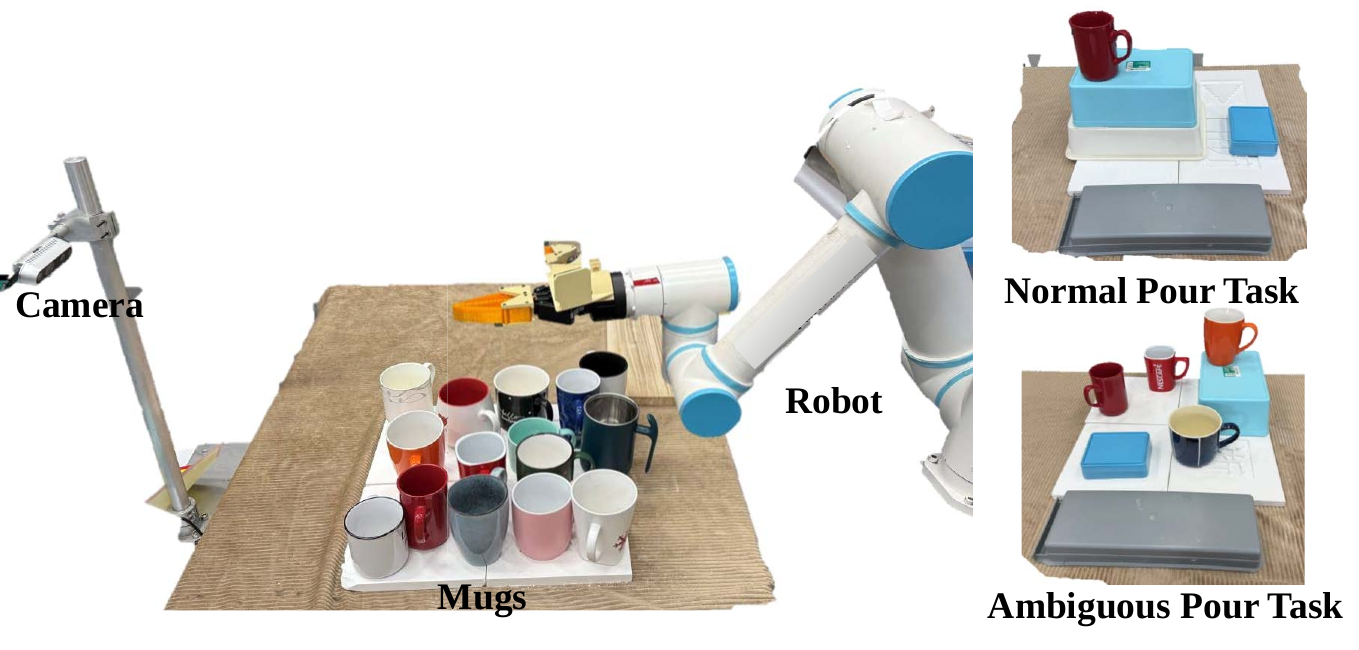}
    \caption{Experimental scene settings. RealSense D435i is used  to capture RGB video streams for perception, and a 6-axis collaborative robotic arm is used for manipulation. In the experiments, over 20 mugs were used for the pouring task. The figure also illustrates the layouts of the \textbf{Normal} and \textbf{Ambiguous} environments. To increase the difficulty, boxes of different colors were randomly placed as obstacles.  }
    \label{fig:scene settings}
\end{figure}

In this section, we investigate whether VL-MP can achieve precise target decision-making and one-shot generalization of robotic trajectories to execute real-world daily manipulation tasks. We design two experimental environments to validate the pouring task, as shown in the figure (Fig. \ref{fig:scene settings}), which illustrates the experimental setup and mugs used for trajectory demonstration. The evaluation metric is the success rate over 10 trials, with success defined as transferring half a cup of water into the target mug without spillage. We test two environments: (i) a normal environment with exactly one target mug, and (ii) an ambiguous environment containing more than three visually similar targets with initially vague task instructions. For each environment, we conduct 10 repeated trials. The table (TABLE \ref{tab:my_label}) shows the final success rates: VL-MP achieves 90\% success in the normal environment and 80\% success in the ambiguous environment. The figure(Fig. \ref{fig:Pouring Performance}) shows the Pouring Performance in normal and ambiguous task scenarios.

\begin{figure}[htbp]
    \centering
    \captionsetup{font={footnotesize }}
    \begin{subfigure}[t]{0.3\linewidth} 
        \centering
        \includegraphics[width=\linewidth]{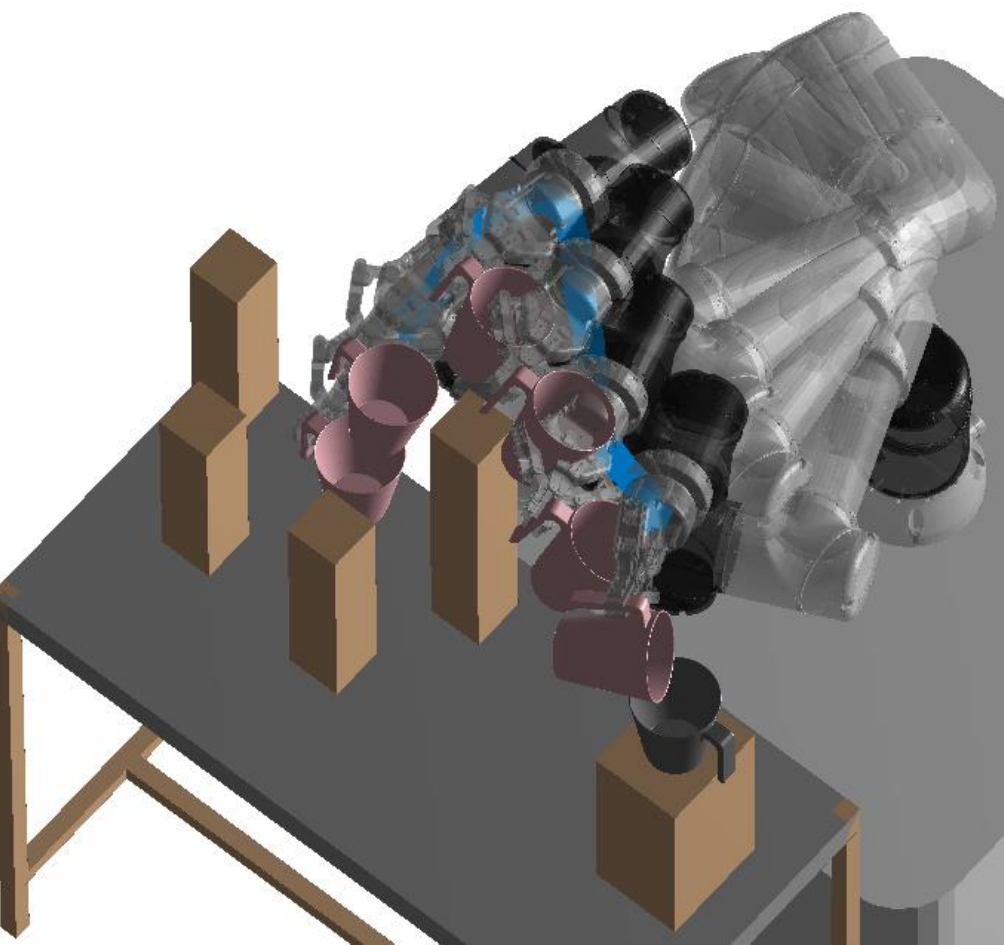}
        \caption{VL-MP(Ours)}
        \label{fig:vl-mp}
    \end{subfigure}
    \hfill
    \begin{subfigure}[t]{0.3\linewidth} 
        \centering
        \includegraphics[width=\linewidth]{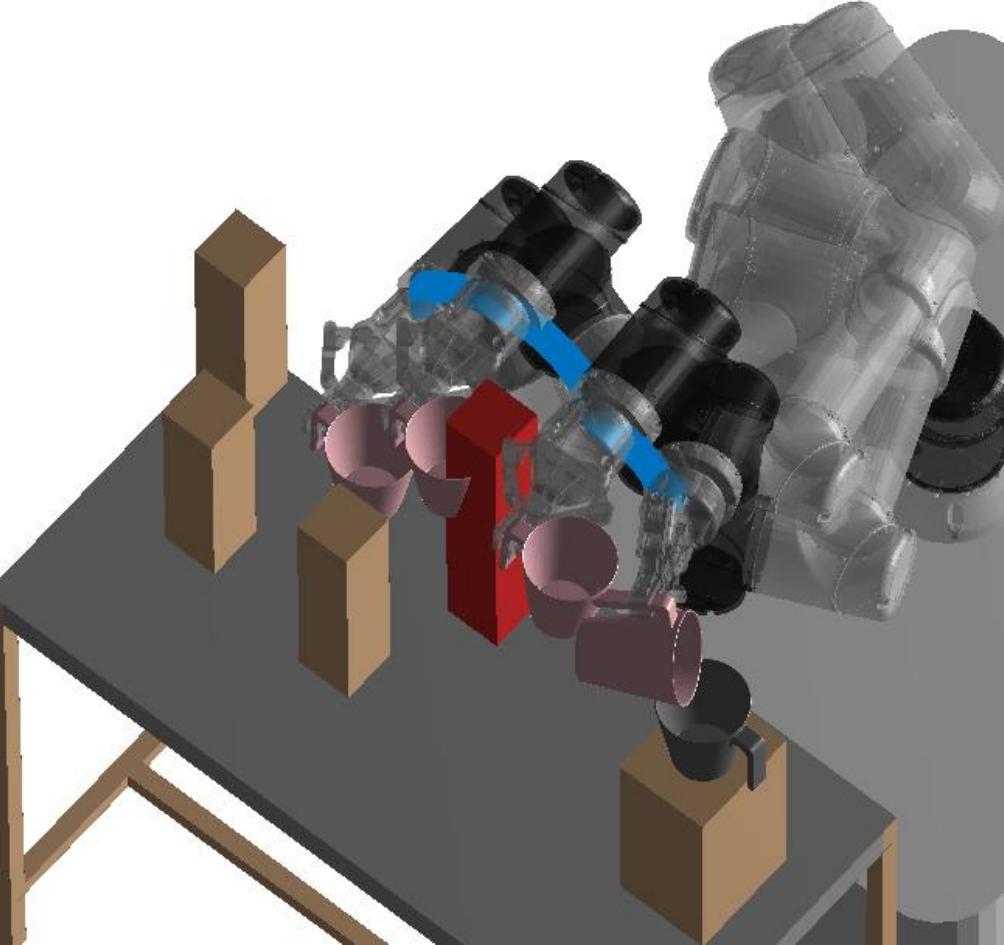}
        \caption{Error breakdown of VLM+Prim and Diffusion}
        \label{fig:error-breakdown}
    \end{subfigure}
    \hfill
    \begin{subfigure}[t]{0.3\linewidth} 
        \centering
        \includegraphics[width=\linewidth]{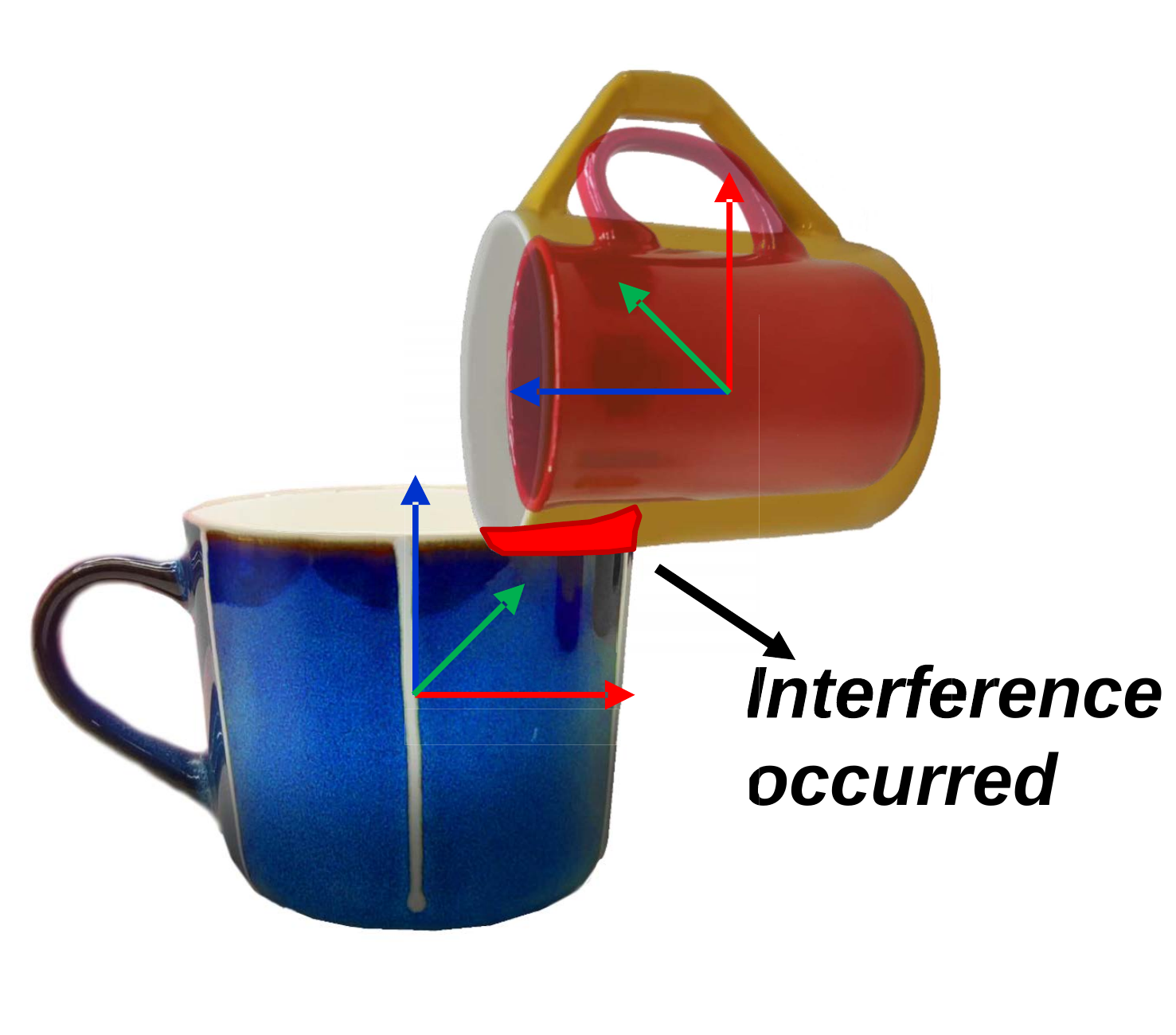}
        \caption{The scene of interference}
        \label{fig:interference-occurred}
    \end{subfigure}
    \caption{Simulation Experiment Results and Error Analysis. (a)Demonstrates the smooth manipulation of VL-MP, which remains robust even when obstacles are added at the initial position. (b)Illustrates that the LLM + Prim and Diffusion method tends to exhibit trajectory divergence at the starting position, leading to collisions. (c)Shows that the LLM + Prim and Diffusion method fails to adapt to intra-class instance shape variations, resulting in mug interference during the final pouring process.}
    \label{fig:simtask-failure-analysis}
\end{figure}

We compare VL-MP with two baselines to validate its effectiveness: (i) VLM+Prim, a variant of Code as Policies \cite{7}, which uses VLMs to parameterize a predefined list of simple primitives and guides the KMP algorithm by transmitting simplified task parameters, and (ii) Diffusion Policy \cite{24}, a generative trajectory imitation learning method that directly learns and outputs skill trajectories without parameter transfer. In the ambiguous task, VL-MP achieves an 80\% success rate, significantly outperforming the baselines (20\% for VLM+Prim and 0\% for Diffusion Policy), demonstrating its ability to leverage VLMs’ logical reasoning and human-robot interaction data to accurately convert decision information into task parameters for execution by the underlying LFE-KMP algorithm. In the normal environment, VL-MP (90\%) also surpasses both baselines (50\%, 40\%), proving that its local feature enhancement improves the precision of trajectory learning  and increases success rates in fine-grained manipulation. These results confirm that VL-MP enables finer-grained skill generalization in ambiguous environments. 
We conducted additional experiments in simulation environments and analyzed failure causes. VL-MP performed robustly in ideal simulation conditions without significant failures(Fig. \ref{fig:vl-mp}). In contrast, the other two methods exhibited frequent collisions near skill initiation points(As shown in Fig. \ref{fig:error-breakdown}). This limitation may be attributed to their lack of local trajectory feature enhancement, leading to trajectory generalization failures in complex environments. 

\begin{table}[H]
    \centering
    \captionsetup{font={footnotesize }}
    
    \caption{Success Rate in Normal and Ambiguous Environments. VL-MP achieved the best performance across all environments. }
    \begin{tabular}{lcc}
\toprule
\textbf{method} & \multicolumn{2}{c}{\textbf{pour}} \\
\cmidrule{2-3}
 & \textbf{normal} & \textbf{ambiguous} \\
\midrule
VLM+Prim & 5/10 & 2/10 \\
Diffusion & 4/10 & 0/10 \\
VL-MP(Ours) & 9/10 & 8/10 \\
\bottomrule
\end{tabular}
    \label{tab:my_label}
\end{table}

\begin{figure}
    \centering
     \captionsetup{font={footnotesize }}
    \includegraphics[width=0.6\linewidth]{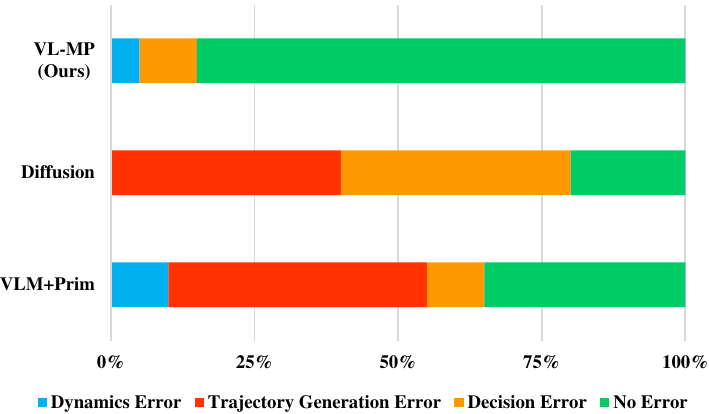}
    \caption{Error breakdown of stages. VL-MP significantly reduces trajectory generation error and decision error.    }
    \label{fig:Analysis of Task Failure}
\end{figure}

We further analyze the experimental results by categorizing errors into three types with ascending statistical priority: (i) Dynamics Error (caused by kinematic computation inaccuracies), (ii) Trajectory Generation Error (due to faulty trajectory synthesis), and (iii) Decision Error (resulting from incorrect task target decisions). As shown in Fig. \ref{fig:Analysis of Task Failure}, Diffusion Policy and VLM+Prim exhibit frequent trajectory generation errors caused by collisions between mug rims. In contrast, VL-MP achieves the lowest overall error rate, with its primary failures attributed to decision errors caused by target occlusion in ambiguous environments. 

\subsection{Shape Preservation of Trajectories Generalization}

In this section, we investigate whether the VL-MP framework can preserve skill trajectory shapes during generalization to learn more challenging precision tasks. Through simulation-based quantitative comparisons, we evaluate its trajectory shape preservation capability using the Handwriting dataset \cite{25}, which comprises handwritten alphabets. We randomly select six samples from the GShape subset for training and generate trajectories to assess shape preservation. To amplify generalization difficulty, the experimental setup significantly alters the relative positional relationships between trajectory start and end points. Comparative analysis is conducted against two imitation learning methods: Kernel Movement Primitives (KMP) \cite{21} and Task-Parameterized Gaussian Mixture Models (TP-GMM) \cite{26}.

To evaluate VL-MP’s trajectory preservation performance, we employ topological similarity ${c_s}$ (as defined in \textit{formula \ref{eq:f13}}) to measure shape preservation during the trajectory generation phase and trajectory smoothness ${\kappa_s}$ (as defined in \textit{formula \ref{eq:f14}}) to assess shape preservation during the execution stage. Smoothness critically affects the preservation of predefined shapes during execution, as smoother trajectories can prevent shape distortion caused by joint impacts, with smaller values indicating higher smoothness. The formulations \(\Delta \xi_{g,t} = \xi_{g,t+1} - \xi_{g,t}\)represent the vectors formed by the \(t\) and \(t+1\) points of the generated trajectory, while \(\Delta \xi_{r,t}\) denotes the corresponding reference trajectory vector. 

\begin{equation}
c_s = \frac{1}{N} \sum_{t=1}^{N-1} \cos \langle \Delta \xi_{g,t}, \Delta \xi_{r,t} \rangle \times 100\%
\label{eq:f13}
\end{equation}

\begin{equation}
\kappa_s = \frac{1}{N} \sum_{t=1}^{N-1} \frac{\Delta \xi_t T_t}{\Delta t^2}
\label{eq:f14}
\end{equation}

\begin{figure}[htbp]
    \centering
    \captionsetup{font={footnotesize }}
    \begin{subfigure}[t]{0.3\linewidth} 
        \centering
        \includegraphics[width=\linewidth]{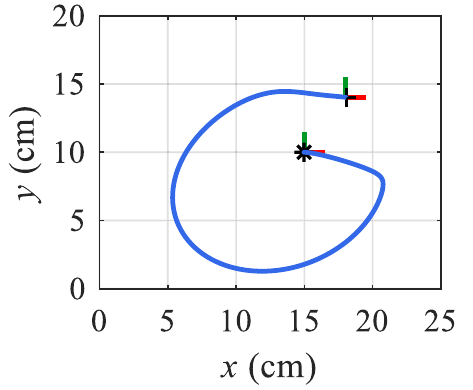}
        \caption{VL-MP(Ours)}
        \label{fig:VL-MP}
    \end{subfigure}
    \hfill
    \begin{subfigure}[t]{0.3\linewidth} 
        \centering
        \includegraphics[width=\linewidth]{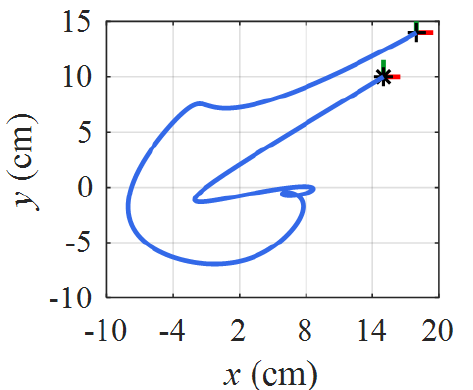}
        \caption{KMP}
        \label{fig:KMP}
    \end{subfigure}
    \hfill
    \begin{subfigure}[t]{0.3\linewidth} 
        \centering
        \includegraphics[width=\linewidth]{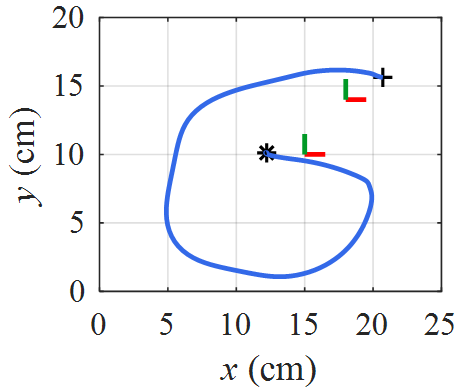}
        \caption{TP-GMM }
    \label{fig:TP-GMM}
    \end{subfigure}
    \caption{Shape preservation performance of trajectory generalization. The "+" symbol and the "*" symbol represent the starting and ending points of the letter "G," respectively. The coordinate symbols correspond to the respective task parameters. VL-MP exhibited the strongest shape preservation capability under disturbances.}
    \label{fig:Comparative-Evaluation} 
\end{figure}

\begin{table}[H]
    \centering
    \captionsetup{font={footnotesize}}
    \caption{Comparison of Shape Preservation Capability. The superior shape preservation capability of VL-MP is demonstrated through the metrics ${c_s}$  ${\kappa_s}$ . }
    \label{tab:my_table_2}
    \begin{tabular}{lcc}
        \toprule
        \textbf{Algorithm} & \textbf{Smoothness} & \textbf{Topological Similarity (\%)} \\
        \midrule
        KMP & 0.4127 & 64.95 \\
        TP-GMM & 0.0911 & 84.89 \\
        SemKMP & 0.0663 & 92.76 \\
        \bottomrule
    \end{tabular}
\end{table}

The experimental results(TABLE \ref{tab:my_table_2}) demonstrate that the VL-MP method outperforms the two baseline algorithms in both trajectory smoothness  ${c_s}$ (0.0663) and topological similarity  ${\kappa_s}$ (92.76\%). The figure(Fig.  \ref{fig:KMP}) shows that the KMP algorithm completely loses the shape characteristics of GShape. This is because KMP encodes and learns the global features of the training set, where the absolute coordinate values of the start and end points heavily influence trajectory generation. As a result, the generated trajectories are confined to the coordinate range of the training samples, and the algorithm forcibly converges to new task parameters, leading to severe shape distortion. Compared to the traditional KMP algorithm, VL-MP achieves a 42.82\% improvement in  ${c_s}$ and an 83.03\% enhancement in  ${\kappa_s}$, indicating that its local feature enhancement significantly improves shape preservation. The figure(Fig. \ref{fig:TP-GMM}) reveals that the TP-GMM algorithm partially retains the shape features of GShape but causes the start and end points of the generated trajectory to deviate drastically from the specified new task parameters, resulting in task failure in real-world scenarios. This occurs because TP-GMM treats all task parameters equally, allowing trajectory features near one parameter to be influenced by others, ultimately leading to significant positional deviations. VL-MP does not suffer from this issue.

\section{Conclusions}

In this work, we present VL-MP, a robotic manipulation framework that bridges VLM and KMP to overcome the existing challenge of achieving fine-grained manipulation and open-set decision-making capabilities simultaneously. The proposed VL-MP effectively transforms high-quality decisions from VLM into task parameters required for KMP's generalization. Additionally, through local feature enhancement, VL-MP addresses the issue of trajectory shape distortion inherent in traditional KMP methods. This results in significant geometric accuracy advantages for fine-grained task generalization.

Comprehensive experimental results demonstrate that VL-MP successfully achieves goal decision-making and fine-grained manipulation in ambiguous environments, making significant strides toward robotic skill learning and generalization. Currently,  VL-MP focuses on the trajectory of the end-effector, though it still maintains strong trajectory generalization and obstacle avoidance capabilities. Future research will focus on VLM-guided whole-arm planning, aiming to achieve more complex and fine-grained manipulations.

\addtolength{\textheight}{-12cm}   




\bibliographystyle{ieeetr}
\bibliography{references}


\end{document}